\documentclass[conference]{IEEEtran}
\IEEEoverridecommandlockouts
\usepackage[hyphens]{url}
\usepackage{cite}
\usepackage{amsmath,amssymb,amsfonts}
\usepackage{algorithmic}
\usepackage{graphicx}
\usepackage{textcomp}
\usepackage{xcolor}
\usepackage{orcidlink}
\usepackage{cite}
\usepackage{siunitx}
\usepackage[format=default]{subcaption}
\usepackage{breakurl}

\def\BibTeX{{\rm B\kern-.05em{\sc i\kern-.025em b}\kern-.08em
    T\kern-.1667em\lower.7ex\hbox{E}\kern-.125emX}}

\begin{document}

\graphicspath{{images/}}

\title{A 1-D CNN inference engine\\for constrained platforms
\thanks{This work is supported by the German Research Foundation (Deutsche Forschungsgemeinschaft, DFG) as part of SPP 2378 (Resilient Worlds: project number 502615015, \textit{Resilient Power-Constrained Embedded Communication Terminals, ResPECT}}
}

\author{
\IEEEauthorblockN{
Ishwar Mudraje\textsuperscript{1 \orcidlink{0009-0003-6870-7862}},
Kai Vogelgesang\textsuperscript{1 \orcidlink{0000-0002-7633-1880}}
Thorsten Herfet\textsuperscript{1 \orcidlink{0000-0002-3746-7638}},
}
\bigskip{}
\IEEEauthorblockA{
\textsuperscript{1} 
Saarland Informatics Campus (SIC),
Germany, 
\{mudraje,vogelgesang,herfet\}@cs.uni-saarland.de}
}

\maketitle

\begin{abstract}
1D-CNNs are used for time series classification in various domains with a high degree of accuracy. 
Most implementations collect the incoming data samples in a buffer before performing inference on it.
On edge devices, which are typically constrained and single-threaded, such an implementation may interfere with time-critical tasks.
One such task is that of sample acquisition.
In this work, we propose an inference scheme that interleaves the convolution operations between sample intervals, which allows us to reduce the inference latency.
Furthermore, our scheme is well-suited for storing data in ring buffers, yielding a small memory footprint.
We demonstrate these improvements by comparing our approach to TFLite's inference method, giving a 10\% reduction in the inference delay while almost halving the memory usage.
Our approach is feasible on common consumer devices, which we show using an AVR-based Arduino board and an ARM-based Arduino board.
\end{abstract}


\section{Introduction}
    % Briefly discussing the applications of time series classification
    The TinyML paradigm focuses on the deployment of machine learning inference and/or training on constrained devices~\cite{abadadeSurveyTinyML2023,ray2022review}.
    Performing inference tasks on IoT edge devices has several advantages, such as reduced latency and preservation of privacy~\cite{deng2020edge}. 
    Traditionally, edge devices offload machine learning inference onto other devices by transmitting raw or processed sensor data to an external device due to the energy and hardware limitations. 
    Over the past decade, developments in hardware, as well as optimization of deep learning training and inference methods, have enabled inference to be performed directly on-device~\cite{sipola2022artificial}.

    \subsection{Edge ML inference}
    IoT devices vary in their capabilities based on the intended application. 
    Modern IoT devices may even be equipped with a dedicated neural processing engine with sufficient memory for on-board inference. 
    FPGA-based frameworks have also been developed for accelerating ML inference on edge devices.

    However, a large number of edge devices consist of microcontrollers (MCU) with a limited amount of computational power and memory. 
    
    Software such as Tensorflow-lite micro, CMSIS-NN and many more have made inference possible on low-resource MCUs through techniques such as post-training quantization of neural networks~\cite{david2021tensorflow, laiCMSIS2018}. 

    \subsection{Time series classification (TSC)}
    The task of time series classification (TSC) involves assigning a label to a sequence of time varying data. 
    Time varying data is ubiquitous, especially in IoT environments where sensors continuously or intermittently generate data. 
    Some well-known applications of TSC in IoT settings are human activity recognition (HAR) and structural health monitoring (SHM)~\cite{ramanujamHAR2021,surucuCondition2023}. 

    TSC requires methods that analyze temporal patterns in multidimensional data. 
    Numerous classical and deep-learning based approaches have been developed for TSC~\cite{faouziTSC2024}. 
    Ensemble methods such as HIVE-COTE 2.0 have proven to be competitive over a variety of datasets~\cite{middlehurst2024bake}. 
    However, ensemble methods are computationally expensive. 
    Lightweight methods such as the recent ROCKET family of methods use random convolutional kernels to achieve comparable accuracy to HC2~\cite{dempster2020rocket}\cite{ruiz2021great}. 
    
    Deep learning architectures developed for TSC have used convolutional layers (CNNs), skip connections (ResNets) and recurrent layers such as long short term memory networks (LSTMs).~\cite{fawazInceptionTime2020} 
    Recent methods have introduced transformer networks with attention layers as well~\cite{chen2021attentionCNN,foumaniDLTime2024}. 
    However, transformer networks are computationally intensive and require parallelism, which is not commonly available on embedded hardware.
    
    \subsection{TSC on resource-constrained IoT devices}
    Deploying TSC models on IoT devices with single-threaded, slow MCUs presents several challenges.
    For instance, a device like an Arduino UNO does not have enough storage capacity to store the weights or enough memory to perform the operations of an SoTA TSC model. 
    For TSC, typically a specified length of data has to be collected before it can be classified.
    This may be prohibitive on devices with limited memory if long sequences need to be collected.
    Moreover, IoT devices execute several tasks such as sensor data acquisition, processing, and communication with other devices.
    Running large TSC methods may also violate timing constraints for acquiring sensor data, if sampling needs to be performed at fixed intervals.
    While interrupt mechanisms can be used to ensure timeliness, interrupt handling will in-turn delay inference tasks.

    \subsection{Motivation and contribution} 
    In several SoTA methods, convolutional layers have been proven to be effective for TSC. 
    For example, InceptionTime uses an ensemble of CNNs while ROCKET uses random convolutional kernels during the feature extraction stage.

    CNNs have the advantage of being sparse in their weights, which makes them suitable for devices with low storage capacity.
    Therefore, deploying small 1D-CNNs is one option to achieve good classification accuracy on highly constrained devices.
    Direct convolution of a kernel with width $M$ on a signal of length $N$ has a computational complexity of $\mathcal{O}(MN)$. 
    For 1D-CNNs, typically $N \gg M$, resulting in a near-linear complexity in signal length.
    Convolutions also have to be performed on multiple channels and for multiple filters.
    Computing the convolutions when the full sequence is available may be too slow on such devices.
    Moreover, storing the input sequence and intermediate outputs may require more memory than available. 

    In practice, MCUs that support multiprocessing or multithreading are used in IoT designs that require neural network inference.
    Using more complex MCUs increases overall development time and design complexity as compared to simpler devices like the AVR series.
    Therefore, inexpensive, but highly resource constrained 8-bit MCUs still form the core of various consumer products~\cite{manners2023Mighty8Bit}.
    A framework that can perform these devices is thus of interest in various consumer applications.
    
    In this paper, we propose an inference method for 1D-CNNs\footnote{Implementation is hosted at \burl{https://git.nt.uni-saarland.de/open-access/cnn-inference-engine}} that interleaves the convolution operations in between sample acquisitions.
    Our method uses ring buffers to store only portions of the signal as required for the convolutional operations.
    This reduces both the inference delay and the memory required for 1D-CNN inference.
    We demonstrate timing measurements to demonstrate the time savings when interleaving convolution operations on an ARM-Cortex MCU.
    We also demonstrate the feasibility of our method for inference on 8-bit single-threaded MCUs.

\section{Related Work}
There is ever-increasing interest in real-time inference on IoT devices.
Research on model compression and acceleration has resulted in both hardware and software implementations of neural network inference for resource-constrained devices.

Some modern sensors can perform signal classification such as activity recognition with on-board AI cores.
In such cases, the TSC task can be offloaded to the sensor, which extracts features using a configurable filter bank and provides the classification result to the MCU~\cite{LSM6DSV16X6axisInertial}.
FPGAs have also been used for inference with neural networks, as direct hardware inference is both faster and is parallelizable~\cite{abtahiFFTEmbedded2018,sanchez1DCNN2018a, zhouAcceleratedFPGA2024}.

In this work, we limit our scope to software approaches that are aimed at consumer grade MCUs, and examine the capabilities that they offer for TSC inference.

The Tensorflow-Lite library provides post-training quantization to both reduce storage requirements and speed up computation. 
Quantization involves converting floating point weights of a neural network into low-precision floating, fixed or even integer quantized forms.
Tensorflow-Lite for Microcontrollers (TFLM) supports a limited range of neural network architectures, which does not include 1-D convolution layers. Instead, 2D convolution layers are used~\cite{david2021tensorflow}.
Similary, the CMSIS-NN provides ARM-specific neural inference kernels for fast inference on the Cortex-M series~\cite{laiCMSIS2018}. 

The SeeDOT and Shiftry frameworks compile ML operations into efficient fixed point weights and operations, resulting in kB-sized models that can fit on tiny IoT devices~\cite{gopinathKBML2019,kumar2020shiftry}. 
Shiftry uses a data-driven fixed-point conversion that adaptively assigns different bit-widths to variables.
Shiftry demonstrated RNN inference on an Arduino UNO.

Alongside model quantization, performing model pruning also results in smaller model architectures with comparable accuracy~\cite{gao2020rethinking}. 
For CNNs,~\textit{Li et al.} demonstrated pruning filters, demonstrating a large decrease in number of parameters~\cite{liPruneCNN2017}.

For CNNs specifically, the convolution theorem also allows for fast implementations of convolution operations:
\textit{Mathieu et al.} demonstrated fast training of CNNs using the fast Fourier transform (FFT)~\cite{mathieuFFTTraining2014}.
FFT-based convolution frameworks have also been developed for MCU targets~\cite{kruppApproxFFT2022}.
However, FFT-based methods are primarily effective for 2D-convolutions.

Apart from inference, platforms such as AIfES and TinyOL can perform on-device training~\cite{renTinyOL2021,wulfertAIfES2024}.

\section{Background}
    \subsection{1D-Convolutional Neural Networks}
    \begin{figure}[t]
    \centering
        \includegraphics{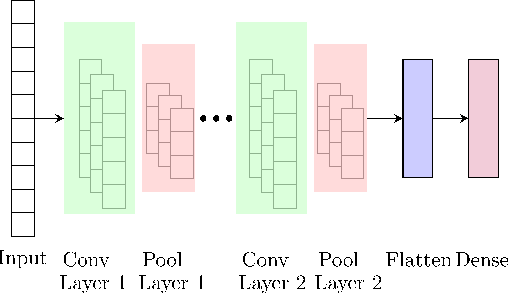}
        \caption{Representation of a 1D-CNN architecture for classification. }
        \label{fig:sample_cnn}
    \end{figure}
    A typical 1D-CNN classifier architecture is represented in Fig.~\ref{fig:sample_cnn}.
    The input to the network is processed by a set of convolutional filter banks to extract features. 
    The operation of a convolutional filter kernel $w$ and bias $b$ of length $M$ on an input signal $x[n]$ of length $N$ is written as:
    \begin{equation}
        \label{eq:1dconv}
        y[n] = \sum_{j=0}^{M} x[n+j] \cdot w[j] + b[j].
    \end{equation}
    \begin{figure}[t]
        \centering
        \includegraphics{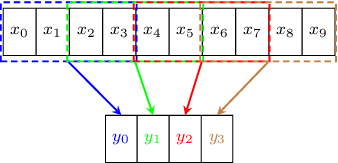}
        \caption{1D convolution operation, visualized as a sliding window operation. Each stride of the kernel is visualized in a different color.}
        \label{fig:sliding_window}
    \end{figure}
    This operation can be visualized as a sliding window operation, as shown in Fig.~\ref{fig:sliding_window}.    
    The stride and padding parameters of convolutional layers respectively dictate the shift of the sliding window in each step and the boundary condition for the convolution operation.
    Each filter is followed by a non-linear activation layer (e.g., ReLU).
    A subsequent pooling layer subsamples the output of the convolutional filters.
    Finally, the output of a series of convolutional layers is passed to a dense layer for the final classification.
    \begin{figure}[t]
    \centering
        \begin{subfigure}[t]{\linewidth}
        \centering
            \includegraphics{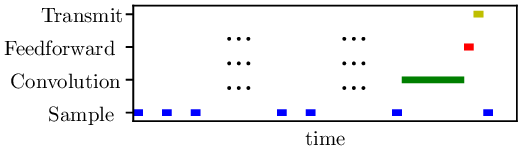}
            \caption{}
            \label{fig:sample_timing_1}
        \end{subfigure}
        \begin{subfigure}[t]{\linewidth}
        \centering
            \includegraphics{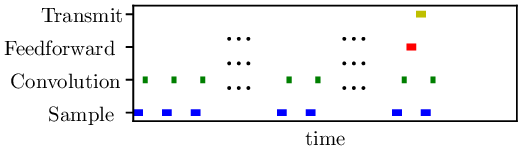}
            \caption{}
            \label{fig:sample_timing_2}
        \end{subfigure}
        \caption{(\subref{fig:sample_timing_1}) Sample timing diagram for the system model. The system waits until the required sequence is collected before performing inference. (\subref{fig:sample_timing_2}) Proposed system model. The convolution phase is interleaved between sample acquisition.}
        \label{fig:sample_timing}
    \end{figure}
    \subsection{System model}
    As IoT devices are heterogeneous, we define a system model that we target for our inference framework.
    We assume an IoT device with a single-threaded MCU connected over a serial interface to a sensor and a communication peripheral to relay data or inference results to an external entity.
    The MCU is assumed to not have any accelerator units for neural network inference.
    We consider in our model that the following tasks are being executed by the MCU:
    \begin{itemize}
        \item Communication over the serial interface with the sensor at a fixed sampling rate to acquire sensor readings, and preprocessing of the sensor data (e.g., filtering).
        \item 1D-CNN inference on a sequence of the sensor data, if necessary.
        The task of inference is split into the feature extraction (convolution) phase and inference phase (feedforward).
        \item Communication of the results of inference or other preprocessing tasks to an external entity.
    \end{itemize}
    
    Fig.~\ref{fig:sample_timing_1} shows an example of the sequential execution of tasks in our system model. 
    In this example, there are two evident problems:\newline
    \textbf{Problem 1}: The computationally intensive convolution task blocks other tasks until completed. 
    Computing the full 1D-CNN inference step after a sequence has been collected also requires more memory.\newline
    \textbf{Problem 2}: The system is idle during the sampling interval. 
    This results in inefficient MCU utilization, especially in the context of inference.\newline
    In the following section, we propose a 1D-CNN inference framework where we aim to alleviate these two problems. 

\section{Methodology}
    \subsection{Decomposing convolutions}
    The whole signal need not be stored for evaluating discrete convolution on it. 
    The overlap-add method has long since been used to break the overall convolution operation into smaller convolutions in a divide-and-conquer strategy.
    A similar divide-and-conquer strategy can also be used for convolution in 1D-CNNs.
    From our example in Fig.~\ref{fig:sliding_window}, for computation of $y_0$, only \{$x_0$, $x_1$, $x_2$, $x_3$\} are used in the inner product.
    For $y_1$, four samples, \{$x_2$, $x_3$, $x_4$, $x_5$\} are needed.
    We can ``forget" $x_0$ and $x_1$ for further evaluation of the convolution.
    For the first stride of the filter, we retain $x_2$ and $x_3$ and append $x_4$ and $x_5$.
    Similarly, for the following stride, we need $x_4$ through $x_7$ and we can forget $x_2$ and $x_3$.
    This follows for inputs to all the convolutional layers of the network.
    These smaller convolution chunks can be interleaved between sample acquisitions, as shown in Fig.~\ref{fig:sample_timing_2}.
    The above semantics align with the operations of a cyclic/ring buffer structure, except for the stride operation, which can be accommodated with a minor modification:

    \begin{figure*}
        \centering
        \begin{subfigure}[t]{0.32\textwidth}
            \centering
            \includegraphics{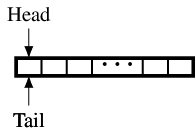}
            \caption{Both \texttt{head} and \texttt{tail} point to the same location: Buffer is empty or full, depending on the previous operation.}
            \label{fig:buffer_empty}
        \end{subfigure}
        \hfill
        \begin{subfigure}[t]{0.32\textwidth}
            \centering
            \includegraphics{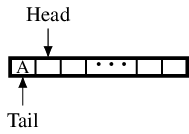}
            \caption{Write operations store the data at the current \texttt{head} location and move \texttt{head} to the next location.}
            \label{fig:buffer_write}
        \end{subfigure}
        \hfill
        \begin{subfigure}[t]{0.32\textwidth}
            \centering
            \includegraphics{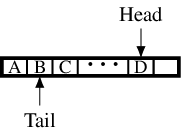}
            \caption{Read operations read from the current \texttt{tail} location and move \texttt{tail} to the next location.}
            \label{fig:buffer_read}
        \end{subfigure}
        \caption{Operations of a ring buffer. \texttt{head} points to the next write location, while \texttt{tail} points to the current read location. The \texttt{stride} operation simply involves moving \texttt{tail} by $s$.}
        \label{fig:buffer}
    \end{figure*}

    Fig.~\ref{fig:buffer} shows the individual operations of a ring buffer adapted to perform a convolution task.
    The buffer has a \texttt{write} method that moves the write pointer (head) by 1.
    Similarly, it has a \texttt{read} method that moves the read pointer (tail) by 1.
    The \texttt{write} and \texttt{read} methods also check if the head and tail coincide and declare if the buffer is full or empty respectively.
    In addition to these methods, a \texttt{stride} method that moves the head by $s$ steps.
    The input ring buffer to a convolutional layer with kernels of size $M$ and stride $s$ then is realized as a ring buffer of size $M$ with the following methods:
    \begin{align*}
        write&: head \leftarrow (head + 1) \mathbin{\%} M,\\
        read&: tail \leftarrow (tail + 1) \mathbin{\%} M,\\
        stride&: head \leftarrow (head + s) \mathbin{\%} M.
    \end{align*}

    \begin{figure*}
        \includegraphics{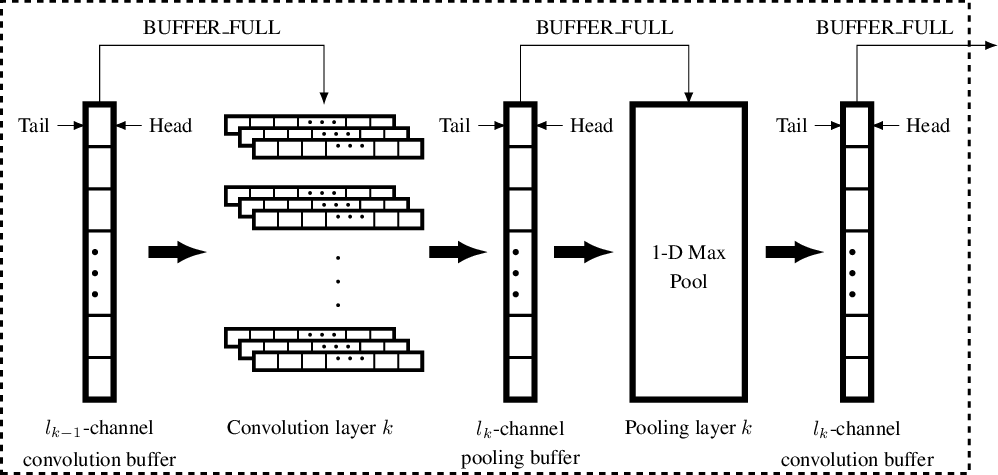}
        \caption{Convolutional block operations of the proposed real-time inference method. Filter and pooling layers are preceded by a buffer. When a buffer is full, the convolution or pooling operation is performed and the result filled into the next buffer.}
        \label{fig:conv_block}
    \end{figure*}

    For a convolutional layer, we require as many ring buffers as there are input channels to the layer.
    We define a convolutional layer as a structure that operates on its input ring buffers.
    Each convolutional layer is provided with the \texttt{step} method that pushes a sample as soon as it is available into its buffer.
    The \texttt{step} methods also checks if the buffer is full and invokes the \texttt{convolve} and \texttt{stride} methods as described in Fig.~\ref{fig:sliding_window}.
    The calculated value is then made available to the next layer, which may then in-turn execute its \texttt{step} method on the available value.
    Note that pooling layers integrate similarly into this framework, executing their \texttt{pool} method -- as opposed to the weighted addition operation of convolution layers -- when their buffers are full.
    Such a convolutional block is visualized in Fig.~\ref{fig:conv_block}.

    Finally, we define a convolutional network structure as a series of convolutional layers.
    Similar to the layers, the convolutional network structure has a \texttt{step} method that invokes the \texttt{step} methods of the individual convolutional and pooling layers. 

    In our system model, the proposed method has the following properties:
    \begin{itemize}
        \item The \texttt{step} method of the 1D-CNN executes after every sample is obtained, resulting in partial computation of the filter layers.
        Thus, \textbf{Problem 2} is alleviated, as the MCU can exploit the previously idle time in between sample acquisitions.
        \item At the end of the duration in which data for classification is collected, the features are already available as they have been computed in steps.
        This addresses the inference latency issue of \textbf{Problem 1}, as only the final classification stage needs to be performed.
        \item The use of ring buffers ensures that only the data required for the current computation is stored in memory, enabling classification of long sequences.
    \end{itemize}
    The proposed method does not speed up the 1D-CNN inference task itself.
    But, the overall time required for inference is reduced through scheduling parts of the convolution task in between sample intervals.
    
    \subsection{Hardware}
    To demonstrate the method, we choose two hardware targets:
    \begin{itemize}
        \item The Arduino Nano BLE 33 (henceforth Nano BLE): Features a 32-bit ARM Cortex-M0 MCU running at \qty{160}{MHz}, along with an on-board inertial measurement unit (IMU).
        TFLM under the Arduino IDE is available for this board, enabling a comparison with our method.
        \item The Arduino Uno/Mega (henceforth AVR): These boards feature an 8-bit AVR Atmega8 MCU running at \qty{16}{MHz}.
        The MCU lacks hardware floating point support but floating point computations are supported through integer arithmetic.
    \end{itemize}
    In both targets, the IMU is connected through I\textsuperscript{2}C to the MCU.
    Firmware development was done using the Arduino framework for Platform IO.
    To obtain timing signals similar to Fig.~\ref{fig:sample_timing}, the GPIO pins of each board are used to signal the beginning and end of each task.
    The signals are captured using an LA5032 logic analyzer by Innomaker.

    \subsection{Evaluation}
    Building on our previous work, we consider a fence intrusion detection scenario where vibration data from an accelerometer is classified into different types of intrusions (e.g., climbing and intentional rattling)~\cite{mudraje2024fids}.
    In our scenario, we fix the sampling rate to \qty{119}{Hz}, as this setting results in a compromise between performance and power consumption of the acceleration sensor.
    Data from the accelerometer is sampled and accumulated for a certain duration of time before being classified.
    We treat the Nano BLE as a single threaded MCU and evaluate the proposed method against TFLM for classifying IMU sensor data.
    As TFLM is unavailable for the AVR devices, we simply measure the corresponding timing as well as the memory usage.

    % Model
    To measure the timings, we use a 4-layer 1D-CNN.
    We use randomly initialized weights as we are only concerned with the execution times and not the classification results themselves.
    3-axis accelerometer measurements of length $\approx$\qty{3.9}{s} each are classified (input of length 460 and width 3). 
    About \qty{5.5}{kB} is required to store the input in raw floating point format.
    Each layer of the 1D-CNN's convolution stage has 8 filters of width 8 and stride 1. 
    The classification stage has two dense layers of width 16, followed by a softmax classification layer.
    The model classifies raw accelerometer data into two classes.
    It has a total of 2338 trainable parameters requiring a storage space of \qty{9.13}{kB} (32-bit floating point).
    
    Our implementation is written entirely in C with statically allocated memory.
    Static allocation is achieved through a model conversion script that converts Tensorflow models into header files that allocate the buffers, convolutional layers and convolutional network structures in a header file.

    The model is created using Tensorflow and ported to both Tensorflow-Lite and our proposed engine using the model conversion script.
    
    \section{Results and discussion}

    \begin{table}
        \centering
        \begin{tabular}{|c|c|}
        \hline
        Task & Execution time (\si{ms})\\
        \hline
        Sample & $1.02 \pm 0.01$\\
        \hline
        Convolution & $502.59 \pm 1.73$\\
        \hline
        Feedforward Step & $1.31 \pm 0.02$\\
        \hline
        Communication & $0.01 \pm 0.00$\\
        \hline
        Latency & $4452.83 \pm 2.42$\\
        \hline
        \end{tabular}
        \caption{Average execution times for different tasks in the system model on the Nano BLE.}
        \label{tab:tf_exec_stats}
    \end{table}

    \begin{figure}
        \centering
        \includegraphics[width=\linewidth]{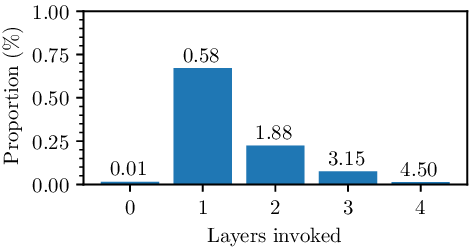}
        \caption{Distribution of execution times for the \texttt{step} method of the proposed 1D-CNN inference engine annotated with average execution times (\si{ms}) for each scenario for the Nano BLE.}
        \label{fig:step_time}
    \end{figure}

    \subsection{Execution time}
    For the 4 layer model, the execution times on the Nano BLE for each of the tasks described in the system model in Fig.~\ref{fig:sample_timing_1}.
    The average execution times are shown in Table~\ref{tab:tf_exec_stats}. 
    As expected, the convolution stage takes the longest to execute.
    For the chosen architecture, the feedforward phase is very short (\qty{1}{ms}).
    
    Further, we define latency as the overall time taken from the start of sample acquisition until the classification result is available.
    The latency also includes the time required for signal acquisition, which is around \qty{3.9}{s} for 460 samples at \qty{119}{Hz}.
    
    For the proposed inference engine, we only show the measurements for the \verb|step| method of the network in Fig.~\ref{fig:step_time}. All other tasks and their execution times remain the same. Four different ranges for execution times are possible, depending on the occupancy status of the input buffers to each convolution layer. The largest values are well within the chosen sampling interval of \qty{8.4}{ms}.
    
    For the Arduino AVR targets, for the chosen model, the average and maximum execution times for the \verb|step| function were measured to be around \qty{12}{ms} and \qty{49}{ms}.
    For these devices, smaller models could be used.
    Stride or number of layers could be varied to further reduce the execution time.
    Another option to reduce execution times is to perform integer quantization, as the 8-bit AVR devices lack a floating point unit.
    However, the measurements clearly demonstrate the limitations of the inference engine when considering slow devices.
    In general, smaller neural network designs have to be considered for such devices.
    Alternatively, lower sampling frequencies could also be used for data acquisition.
    
    \subsection{Memory usage}
    For the proposed inference engine, all objects are statically allocated. 
    With the chosen architecture, the overall RAM usage reported by the compiler for the Nano BLE is around~\qty{45}{kB}.
    On the other hand, for the Tensorflow version, at least~\qty{85}{kB} was reported during compilation. 
    In both cases, the reported memory usage is not exclusive to the inference task and includes RAM usage for other tasks.
    However, the reduction in memory usage is due to the proposed inference method, as other tasks remain the same.

    For the AVR, the RAM usage is reported to be slightly higher than~\qty{2}{kB}.
    The low memory usage means the inference engine can be used on devices smaller than the Mega 2560, especially if smaller models are used.
    Such a model was demonstrated in our previous work~\cite{mudraje2024fids}.
    
    \section{Conclusion and future work}
    The proposed 1D-CNN inference engine offers a lower-latency inference result while also reducing memory usage on single-threaded low-resource MCU targets.
    The inference engine interleaves convolution tasks in between sample acquisitions in order to reduce the overall time required to obtain a classification result.
    Further, a reduction in memory required for inference was demonstrated, making the inference engine feasible for inference on 8-bit devices. 

    In this work, we do not consider post-training optimization methods such as quantization or model pruning.
    Our inference framework in combination with the powerful quantization features offered by libraries such as Tensorflow and AIfES, actual inference time as well as the memory requirements can be further reduced.

\bibliographystyle{IEEEtran}
\bibliography{cnn_inference}

\end{document}